\documentclass{article}

\usepackage{arxiv}

\usepackage[utf8]{inputenc} 
\usepackage[T1]{fontenc}    
\usepackage{hyperref}       
\usepackage{url}            
\usepackage{booktabs}       
\usepackage{amsfonts}       
\usepackage{nicefrac}       
\usepackage{microtype}      
\usepackage{lipsum}		
\usepackage{graphicx}
\usepackage{natbib}
\usepackage{doi}
\usepackage{mathtools}
\usepackage{multirow}
\usepackage{adjustbox}
\newcommand{\e}{\begin{equation}}
\newcommand{\ee}{\end{equation}}
\def\b{\ensuremath\boldsymbol}

\newtheorem{theorem}{Theorem}
\newtheorem{proposition}{Proposition}

\newtheorem{corollary}{Corollary}

\newtheorem{remark}{Remark}
\newtheorem{algorithm}{Algorithm}

\title{Nonparametric Functional Analysis of Generalized Linear Models Under Nonlinear Constraints}

\author{ \href{https://orcid.org/0000-0002-3289-8922}{\hspace{1mm}Kali P.~Chowdhury}\thanks{I would especially like to acknowledge Laura Smith, Weining Shen and Knut Solna for their invaluable insights into the various theorems and their applications. I would also like to thank the University of California, Irvine's Machine Learning Lab for their generous access to the data.} \\
Johns Hopkins University\\
	\texttt{dukechowdhury@icloud.com} \\
}

\date{}

\renewcommand{\shorttitle}{\textit{arXiv}}

\hypersetup{
	pdftitle={Nonparametric Functional Analysis of Generalized Linear Models Under Nonlinear Constraints},
	pdfsubject={},
	pdfauthor={K. P. Chowdhury},
	pdfkeywords={Unbalanced Data; MCMC; Artificial Intelligence; Machine Learning; Nonparametric Regression and Categorical Data Analysis, Model Diagnostics.},
}

\begin{document}
	\maketitle
	
	\begin{abstract}
This article introduces an unified nonparametric methodology for Generalized Linear Models using a unique measures given any model specification. Requiring minimal assumptions, it extends recently published work and generalizes it. If the underlying data generating process is asymmetric, it gives uniformly better prediction and inference performance over the parametric formulation. Furthermore, it introduces a new classification statistic utilizing which I show that overall, it has better model fit, inference and classification performance than the parametric version, and the difference in performance is statistically significant especially if the data generating process is asymmetric. In addition, the methodology can be used to perform model diagnostics for any model specification. This is a highly useful result, and it extends existing work for categorical model diagnostics broadly across the sciences. The mathematical results also highlight important new findings regarding the interplay of statistical significance and scientific significance. Finally, the methodology is applied to various real-world datasets to show that it may outperform widely used existing models, including Random Forests and Deep Neural Networks with very few iterations. 
	\end{abstract}


\section{Introduction} \label{sec:intro}
Binary outcome models continue to be relevant for Artificial Intelligence (AI) and Machine Learning (ML) applications in the sciences \cite{hu2020identification,CHOWDHURY2021101112}, as they serve as the building blocks for various multinomial extensions \cite{murad2003small}. Thus, their importance for AI and ML applications is also widely recognized \cite{li2018measuring}. As such, improvements of these foundational methods remain important for the greater scientific community. In addition, given their ubiquity to the sciences, any proposed improvement over existing methods should retain and be equivalent to the exisitng frameworks in a logical and consistent way if the data support them. This article presents such a framework rooted in Real and Functional Analysis with numerous advatages in regards to Model fit, Inference, and Prediction (MIP) over existing methods. However, before highlighting these contributions, a broad comparison to existing frameworks is needed given the popularity of such models.

In particular, existing methods whether applied in a latent variable (LV) or binary outcome (BO) specification remains field specific. For example, in Econometrics LV models have been used to understand behavior of the average individual within a population \cite{greene2003econometric}, for calculating propensity scores for causal interpretation and program evaluation \cite{imbens2015causal}, and also to understand heterogeneity through finite and infinite mixture distributions \cite{andrews2002hierarchical}. In Biomedical Sciences \cite{zhang2017predicting}, and in the Physical Sciences \cite{hattab2018case} there is also an extensive history of each formulation, and this is true of many fields across the sciences.

Indeed, from \cite{CHOWDHURY2021101112}, it is also apparent that the underlying assumptions of BO vs. LV models may be distinctly different. Thus, it may be difficult to reconcile divergence in MIP performances between their applications. Further complexities arise if the data are unbalanced as then the assumptions of popular models such as the Logistic (logit) or Probit (probit) models need not hold. Thus unsurprisingly, it is well established that the parameter estimates in these models, in either BO or LV models are susceptible to bias and inconsistency \cite{simonoff1998logistic, abramson2000parameter, maity2018bias}. To overcome some of these issues, \cite{CHOWDHURY2021101112} presented a parametric extension of the current Genralized Linear Model (GLM) framework. The work had multiple contributions. Applied in the logit formulation it gave equivalent performance to existing GLMs such as the logit if the data supported their assumptions, but could give better MIP results if they were violated. The methodology also gave results better or equivalent to popular AI methods such as Artificial Neural Network (ANN) under a wide range of circumstances without loss of interpretability of parameter estimates. In addition, the methodology introduced a large-sample diagnostic test which could be used to improve existing AI methods. As such, the work presented a better baseline against which popular AI and ML methods could be compared with better coverage probabilities than existing widely used methodologies such as the maximum likelihood (mle) logit regression.

Further, the functional specification was shown to be highly flexible, since the link condition automatically adjusts to violations of the link constraint. This is because the link constraint held conditionally for all observations. However, the underlying probability of success in its formulation was assumed to be parametric in design. Thus, despite a flexible link function, the estimation of the model when the distribution on the underlying latent error differs from the parametric specification can potentially lead to minor technicalities. As such, this paper adds to this parametric version presented in \cite{CHOWDHURY2021101112} using an entirely novel nonparametric application termed Latent Adaptive Hierarchical EM Like algorithm. Though the parametric version remains relevant for inference, especially if the underlying data generating process (DGP) is symmetric, the nonparametric application can improve upon it for classification purposes in training datasets and can outperform it in test datasets if the underlying DGP is asymmetric. The simulation studies also paint a more nuanced picture of when the parametric or nonparametric versions are more (or less) useful in comparison to existing AI and ML models or GLMs. 

Thus, this article presents several meaningful extensions of the extant literature that spans all three aspects of MIP. In particular, for convergence results, in simulation studies I show that the convergence of the nonparametric application takes longer if the underlying DGP is asymmetric, but has very similar performance to the parametric setting if the DGP is symmetric. For prediction, if the DGP is symmetric it can outperform the parametric version for training datasets but has very similar prediction performances in test datasets. Furthermore, for symmetric DGPs it has largely equivalent or at best nominally better inference performance to the parametric methodology. However, if the data are asymmetric it has better overall prediction and inference performance to the existing parametric version. To better comprare classification performance among the various methodologies considered, I also introduce a new ROC-Statistic based statistical test based on \cite{chowdhury2019supervised}. This large-sample test allows model performance comparison to understand if divergences are statistically different. 

In addition, in one of the more useful applications of the methodology, a separate and novel large-sample test is introduced, which allows us to test whether any particular parametric assumption on the DGP actually fit the observed data, without any a priori assumption on the DGP itself. As such it extends \cite{liu2018residuals} for a broader understanding of model diagnostics for data analysis. Thus, in what follows I first present the mathematical foundations of the methodology in Section \ref{sec:meth}, postponing all technical proofs to the supplementary materials. Extensive simulation studies are performed in Section \ref{sec: mon2} followed by multiple applications of the methodology to real-world datasets in Section \ref{sec: emp}. I then discuss the findings in Section \ref{sec:discch3} where I also discuss the implications of the mathematical results on our continuing discussion of statistical significance and scientific significance. Finally I end with some concluding thoughts in Section \ref{sec:conc}.

\section{Methodology}\label{sec:meth}
Following the notation of \cite{agresti2003categorical} note that a GLM expands ordinary regressions to atypical response distributions and modeling functions. It is identified by three components, the random component for the response $ \b{y}, $ a systematic component that outlines how the explanatory variables are related to the random components, and a link function. The link function specifies how a function of $ E(\b y) $ (please note that for notational simplicity I suppress the dependence on $ X $ unless otherwise states) relates to the systematic component. The random components of $ \b y $ are considered independent and identically distributed (i.i.d.)\footnote{Henceforth, uppercase letters indicate matrices and bolded letters indicate vectors.}. Thus, consider a $n \times 1$ outcome variable $ \b{y} $ which is related to a $ n \times (k+1) $ set of explanatory variables, $ X=(\b{1}, \b{x_1},..., \b{x_k}) $, with $ \b{x_k} $ and $ \b{1} $ each $ n \times 1 $, through a continuous, bounded, real valued function $ c(X) $ of the same dimensions, namely $ n \times (k+1) $. The usual $ (k+1) \times 1 $ parameters of interest are $ \b \beta =\{\beta_0,...,\beta_{k}\}  $, where each $ \{\beta_k\} $ can be a vector.
%
%
%
%
%

To retain the current models should the data support them, I follow the more general framework as in \cite{CHOWDHURY2021101112}, and for completeness, restate the latent variable formulation as considered in \cite{albert1993bayesian} here as well. Let $ \b y^* $ be a latent or unobserved continuous random variable. Then an index function model for binary outcome gives the GLM, \e
\b y^*  = c(X) \b{\beta} + \b{\epsilon},
\ee
where \e
y_i = \begin{cases}
	1\ if\ y_i^* \ge 0,\\
	0\ if\ y_i^* < 0, 
\end{cases}
\ee
with the threshold 0 being a normalization. The two approaches have their strengths and weaknesses, and the purpose of this contribution is to better align the advantages of both models in a rigorous fashion using some of the findings of  \cite{CHOWDHURY2021101112}. For example, Albert and Chib [Ibid] clearly point out that in the binary regression case in the Frequentist interpretation any observed error can only take two values, either 1 or 0. On the other hand, in the latent variable formulation the existence of such an unobserved variable $ \b y^* $ is not guaranteed in the current formulation. The reasoning why the application still holds especially in the symmetric DGP case is due to \cite{tanner1987calculation} as in Albert and Chib [Ibid], $ \b y^* $ is integrated out. In making such an assumption it is also necessary to fix the variance of the latent distribution for identification purposes. An approach which under most circumstances would be considered restrictive.

The proposed methodology has two main goals currently. It first seeks of a way to incoporate the strengths of both the latent variable and the binary regressions in a mathematically rigorous way. Secondly, it seeks to overcome restrictive assumptions on the latent distribution such as having a constant variance or assuming a particular distribution on the probability of success. Thus, below first I outline the mathematical results for the latent variable formulation using signed measures which has distinct advantages over the current latent variable formulation. For a nontechnical discussion of this framework I refer the reader to the supplementaty materials for this paper\footnote{The proofs of the results are also postponed to the supplementary materials.}. Then I discuss the large sample tests for model diagnostic and then briefly highlight the asymptotic properties of ARS.

\subsection{Mathematical Results}\label{math2}

Below, I first present the mathematical foundations for the discussions above\footnote{Some immediately relevant definitions may be found in the supplementary materials for this paper. I kindly refer the reader to any graduate level Analysis book for the remainder of the definitions.}. For a discussion on when the latent variable and binomial regression formulations are equivalent, I refer the reader to \cite{CHOWDHURY2021101112}\footnote{For the current formulation I assume that the formulations are equivalent.}. Thus, I take the results of Chowdhury [Ibid] and the insights from the discussions above to present a more coherent framework that unifies the methodologies in a mathematically rigorous way.



To that end note that it is well established from analytic theory that the restriction of a measure space to a subset of the measureable space is also measureable. However, for the current GLM construction we want to focus on a particular subset, that of the link function. Note that from previous discussions we know that by construction a link function relates the systematic components to the mean of an observation in a specified manner, i.e., $ \eta_i = \lambda(\b x_i,\b\beta) $.
The novelty of this methodology is in considering a signed measure over the $ \sigma-algebra$ defined on the support of the link function.
\subsubsection{Mathematical Foundations of the Proposed Methodology}
Accordingly, first note that the restriction of a measure space to a subspace of the $ \sigma-algebra $ is itself a measure space. For the purpose at hand we seek to restrict our attention to the subspace of the sample space that ensures that the link condition holds pointwise. Thus, consider the measure space $ (X, \Sigma_0, \nu_0) $ restricted to a subspace of $ X $ for which the link condition holds for any particular $ \b\beta $. Then from elementary analysis we know that $ (X, \Sigma_{0|\b\lambda(X,\b\beta)} = \Sigma,\nu_{0|\b\lambda(X,\b\beta)} = \nu) $ is also a measure space. The results below outline this in a more rigorous way.

\begin{proposition}\label{prop1}
	For the measureable space $ (\lambda(X,\b\beta), \Sigma) $ There exists a signed measure  $ \nu $ such that, \e y^* = \begin{cases}
		1\ if\ \lambda \ge 0\\
		0\ if\ \lambda < 0,
	\end{cases}
	\ee
	where WLOG $ \lambda \in [-\infty, \infty) $.
\end{proposition}

Above and for the remainder of this manuscript for notational simplicity I employ $ \lambda $ to represent $ \lambda(X, \b\beta) $. The preceding proposition established the existence result. The forthcoming proposition establishes the uniqueness of this construction for the finite case.
\begin{proposition}\label{prop2}
	For the measureable space $ (\lambda, \Sigma) $ there exists an unique decompostion of  the signed, finite measure  $ \nu$ as a function of two positive measures $ \nu^+ $ and $ \nu^- $ such that,
	\e \nu = \nu^+ - \nu^-\ where, \ee
	\e y_i^* = \begin{cases}
		1\ if\ \lambda \ge 0\\
		0\ if\ \lambda < 0,
	\end{cases}
	\ee
	and $ \lambda \in (-\infty, \infty) $.
\end{proposition}
%
%
%
%
\begin{proposition}\label{prop3}
	Let $ (\lambda, \Sigma, \nu) $ be a measure space as above and $ \nu $ a finite signed measure on it. Then, \e |\nu|(\Sigma) = \bar{\nu}^+(S^+)+ \bar{\nu}^-(S^-)\ee is a probability measure, where $ \{\bar{\nu}^+, \bar{\nu}^-\} $ are positive measures and $ S^+ $ is positive, but $ S^- $ is negative w.r.t. the signed measure $ \nu $.   
\end{proposition}
%
%
\noindent Before going to the next result we need another consequence of an infinite measure space with some collection of measureable sets which are finite.
\begin{theorem}\label{theo0}
Let $ (X, \Sigma, \mu) $ be any measure space with $ \{E_k\}^{\infty}_{k=1} \subseteq A \subset \Sigma $ a collection of measureable sets for which $ \mu(A) = \infty $. Then there exists some $ E_j $ where j is a countable collection of some k, such that $ \mu(\cup_jE_j) < \infty $.
\end{theorem}

\noindent The existence of this result is important for dealing with signed measures, which can take one of the nonfinite values which is not $ \sigma-finite$. Therefore, utilizing this result we can now show one of the more important results and corollaries of Theorem \ref{theo0}, which will be important for the construction of finite or $ \sigma-finite$ measure spaces for all GLMs. It is detailed below.
\begin{proposition}\label{prop4}
Let $ \left(\lambda, \Sigma\right) $ be a measureble space with $ \mu $ a measure which is neither finite or $ \sigma-finite $ such that $ \nu(\Sigma) = \infty $ and let $ \left(\lambda, \Sigma, | \bar{\nu} |\right) $ be a $ \sigma-finite $ measure space. Then if the signed measure $ \nu $ takes one of the values of $ \{-\infty, \infty\} $ then either $ y=1 $ or $ y=0 $ for every observation w.r.t. the $ \sigma-finite $ measure.
\end{proposition}
This result has some very important consequences on the usual regression analysis widely used in the sciences. For example, we may now consider a finite measure such that for a measureable set $ E \in \Sigma $, \e
| \bar{\nu} | = | \nu |\delta_{E \cap S^+},\ where\ \delta_{E \cap S^+}=\begin{cases}
1\ if\ E \in S^+\\
0\ o.w. 
\end{cases}
\ee

The usefulness of the result follows from the unique Jordan Decomposition of a signed measure. If $ f $ is a Lebesgue integrable function then existing results from analysis can be used through the translation invariance property of the measure, to find the unique functional specification as demonstrated in a forthcoming corollary. The astute reader no doubt realizes that in the case that the signed measure takes one of the $ \{-\infty, \infty\} $ values, over a measure space which is neither finite nor $ \sigma-finite$, then we may have information loss. The resulting reformulation to the restricted measure space given in Theorem \ref{theo0} and Proposition \ref{prop4}, can be overcome to address this information loss concern, and the following proposition addresses this\footnote{The relevant definitions can be found in the supplementary materials}. Thus, the Hahn-Banach Theorem can be used to define a linear functional over all integrable functions in $ L^p(\lambda, \Sigma,  |\bar{\nu}|) $ with $ 1 \le p < \infty $.
%

%
%
\begin{proposition} \label{prop5}
Consider a Hahn-Decomposition of the measure space $ (\lambda, \Sigma, \nu) $ into $ \{S^+, S^-\} $ as defined before, where the signed-measure $ \nu $ WLOG takes the value of $ -\infty $ but is not $ \sigma- $finite. Then there exists a linear functional $ \mathcal{L} $ which extends any measure $ \nu^+ $ over $ S^+ $ to all of $ L^p(Q, |\bar{\nu}|) $, with $ |\bar{\nu}| $ as in Proposition \ref{prop4}, and Q is the measureable space $  (\lambda, \Sigma)  $ with $ 1 \le p < \infty $.
\end{proposition}
%
%
%

Using these results then we have some useful existing results from Real Analysis which can be restated for the specific purpose at hand.
\begin{corollary}\label{coro1}
Let $ \left(\lambda, \Sigma, | \bar{\nu} |\right) $ be a $ \sigma- $finite measure space, where $\bar{| \nu |} $ is defined as in proposition \ref{prop4}.  Let $ \{f_n\} $ be a sequence of bounded Lebesgue measureable functions finite a.e. that converges p.w. a.e. on the set $ E \in \Sigma\setminus S^- $ to f which is also finite a.e. on E. Then, \e
\{f_n\} \rightarrow f,
\ee 
in measure.
\end{corollary}
%
%
\begin{corollary}\label{coro2}
Let $ \left(\lambda, \Sigma, | \nu |\right) $ be a $ \sigma- $finite measure space, where $ | \nu | $ is defined as in proposition \ref{prop3}.  Let $ \{f_n\} $ be a sequence of bounded Lebesgue measureable functions finite a.e. that converges p.w. a.e. on the set $ E \in \Sigma $ to f which is also finite a.e. on E. Then, \e
\{f_n\} \rightarrow f,
\ee 
in measure.
\end{corollary}
%
%
These results point to convergence in probability of the parameters in the current formulation. While convergence in probability is useful, below a stronger result is given in Theorem \ref{theo2}. Furthermore, these results also have several nonintuitive  applications to non-binary analysis and the remarks below highlight some of them. 
\begin{remark}
First, note that the decomposition above is unique, and as such the existence of a latent variable implies the existence of an unique pair of positive measures that can represent it.
\end{remark}

\begin{remark}
If the signed measure takes one of the values of $ \{-\infty, \infty\} $ but is not $ \sigma- $finite, we may represent any continuous generalized linear model in one of the outcomes with possibly a linear transformation that ensures all observed $ (\b y,\b x) $ as a function of $\b \beta $ are positive or negative (WLOG). As such the traditional regression formulation can similarly be improved using the link-constraint condition holding for each observation!
\end{remark}


\begin{remark}
That a $ \sigma- $finite signed measure can be extended to a complete measure space $ (\lambda, \Sigma,  \nu ) $ with $ \nu $ a restriction of the outermeasure on $ \Sigma $ follows from elementary analysis results. In addition, while the traditional formulation assumes a symmetric distribution around the mean (for example $ N(\lambda,1) $), the current formulation allows far more flexibility. This is because, instead of fixing the variance of an unimodal symmetric distriution we may instead fix the value based on a probability as a function of $ \lambda $. As such, the latent variable formulation can take a symmetric or asymmetric distributional form around 0, and thus the probabilities of success do not necessarily have to approach either 0 or 1 at the same rate.
\end{remark}

\begin{remark}
That a finite signed measure can be extended to a complete measure space $ (\lambda, \Sigma, | \nu |) $ with $| \nu |$ a restriction of the outermeasure on $ \Sigma $ follows from elementary analysis results. In addition, while the traditional formulation assumes a symmetric distribution around the mean (for example $ N(\lambda,1) $), the current formulation allows far more flexibility. This is because, instead of fixing the variance of an unimodal symmetric distriution we may instead fix the value based on a probability as a function of $ \lambda $. As such, the latent variable formulation can take a symmetric or asymmetric distributional form around 0, and thus the probabilities of success do not necessarily have to approach either 0 or 1 at the same rate.
\end{remark}
\noindent In the forthcoming, I elaborate on the convergence properties of the methodology. However, first I give some foundational results for the uniqueness of the link constraint.
\begin{theorem}\label{theo1}
Let $ (\lambda, \Sigma) $ be a measureable space. Then there is an unique solution to any link modification problem, where the link constraint holds with equality in the Generalized Linear Model Framework for some $ \alpha^* \in R\setminus \{-\infty,\infty\} $, given $ \hat{F}_i $ $ \notin \{0,1\} $, $ X \notin \{0, \infty, -\infty\} $ element wise for each $ i  \in \{1,...,n\}$ and $\b \beta_j \notin \{\infty, -\infty\}$ with $ j \in \{1,...,(k+1)\} $.
\end{theorem}
%

%
%
%
%
%
%
%
%
%
\noindent I now discuss the almost sure convergence property of this methodology.
\begin{theorem}\label{theo2}
Given $ \alpha^* \in R\setminus \{-\infty,\infty\} $, and $ \hat{F}_i $ $ \notin \{0,1\} $, $ X  \notin \{0, \infty, -\infty\} $ elementwise for each $ i  \in \{1,...,n\}$ and $\b \beta_j \notin \{\infty, -\infty\}$ with $ j \in \{1,...,(J+1)\} $ subject to the link constraint holding for each observation, \begin{equation}
	\hat{\beta} \xrightarrow[]{a.s.} \beta.
\end{equation}
\end{theorem}
\noindent This result has some ready extensions to the $ L^p $ spaces and the result below highlights one of those results.
\begin{corollary}\label{coro3}
Under the conditions of Theorem \ref{theo2} we have that for $ 1\le p < \infty $, \e \{g_n\} \xrightarrow[]{a.s.} p(\beta|y). \ee
\end{corollary}
%
%
One of the more useful results of the methodology is that the latent variable nonparametric distributional assumptions do not need any restrictions on the variance parameter. The next corollary puts this in more concrete terms. Thus, we can assert the following regarding the variance of the nonparametric latent variable formulation.
\begin{corollary}
Given $ \alpha^* \in R\setminus \{-\infty,\infty\} $, and $ X  \notin \{0, \infty, -\infty\} $ for each $ i  \in \{1,...,n\}$ and $ \b \beta_j \notin \{\infty, -\infty\}$ with $ j \in \{1,...,(k+1)\} $, the variance of the latent variable distribution $ y^* $ need not be fixed for identification subject to the link constraint holding for each observation.
\end{corollary}
%
\noindent These results are fundamental to the existence and uniqueness of the methodology based on signed measures. However, in order to apply it, an equally important aspect is the estimation procedure. Next this is elaborated.
\subsection{Nonparametric Latent Adaptive Hierarchical EM Like (LAHEML) Algorithm}

The mathematical foundations above may be used to implement an extremely versatile almost sure convergent methodology for the parameters of interest, using a Bayesian Hierarchical MCMC framework through data augmentation. I refer to the estimation methodology as Latent Adaptive Hierarchical EM Like Algorithm (LAHEML), which is nevertheless more general than the EM algorithm. This is because following \cite{CHOWDHURY2021101112}, only ergodicity and aperiodicity are required for the convergence results to hold. In fact, the mathematical foundations ensure that the adaptive algorithm may also allow for both model fit and model selection to be performed at the same time. While specifc applications of the algorithm can be found in the supplemetary materials in both penalized and unpenalized formulations, here I present the general algorithm.

\begin{algorithm}	
\hrule
\vspace{.1in}
\textbf{LAHEML Algorithm}
\vspace{.1in}
\hrule

\vspace{.1in}

\textbf{Require:} Starting values for parameters $ \b{\beta}^{(j)} $. Functional specification $ f(\b y, X) $.
\begin{enumerate}
	\item Subject to the link constraint holding for each observation, compute the truncated signed measures as a function of $ \kappa $.
	\item Compute the link constraint parameter distribution for $ \alpha^{(j+1)*} $.
	\item  Perform an MH step to get $\beta^{(j+1)} $ as a function of $\alpha^{(j+1)*} $ for the relevant likelihood through data augmentation such that, $ \beta^{(j)} \leftarrow\beta^{(j+1)}$.
		\item Iterate to completion. 
	\end{enumerate}
\end{algorithm}
\hrule
\vspace{.2in}
The actual implementation of the algorithm is rather involved because of the need to consider the likelihood principle in the various steps. As such, specific applications in unpenalized and penalized formulations may be found in the supplementary materials. In both applications the mathematical construct ensures that $ \{\b{\beta}^{(j)}\} $ converges to its true distribution given the data $ \b y $. Since $ \b{\alpha^*} $ is a function of $ \b{\beta} $, the convergence results hold for its distribution as well. In particular, the bias of the nonparametric density estimates are corrected by ensuring the link condition holds for each observation. In the implementation of the LAHEML algorithm, it is also worthwhile considering that the cutoff points of $ \kappa \in (0, 1)$ a probability, may also be a parameter to be estimated here. Since the Jordan decomposition remains valid, such a model specification of the methodology should be especially relevant for asymmetric DGPs. This is pursued in the nonparametric simulation datasets, where the cutoff was based on the observed distributions of successes and failures and not necessarily fixed at the median for $ f(\alpha^*|\b\beta,\b y^*,\b y) $. The results of the simulation studies can be found in Section \ref{sec: mon2}. 

A penalized application was also considered for LAHEML using the Bayesian Adaptive Lasso as in \cite{leng2014bayesian}. This penalized version of the methodology is contingent on a different prior specification than in the unpenalized case. Specifically, in this case the prior is given by, 
\e \pi(\b{\beta}|\sigma^2) = \prod_{k=1}^p\frac{\lambda_k}{2\sqrt{\sigma^2}}e^{\lambda_k|\beta_k|\sqrt{\sigma^2}}. \ee
For the highlighted application we may move forward with either a Hierarchical formulation or an empirical bayes application as given in Leng et al. [Ibid] and I follow a Hierarchical methodology accordingly. This is because it also requires the estimation of the prior hyperparameters on the shrinkage parameters, which can no longer be considered uninformative. \cite{leng2014bayesian} consider a gamma prior on the $ \lambda_k $'s and I also follow this same formulation. Thus, the prior on the shrinkage parameters can be given by \e
\pi(\lambda_k^2) = \frac{\delta^r}{\Gamma(r)}(\lambda_k^2)^{r-1}e^{-\delta\lambda_k^2}.
\ee
Further note that \cite{lehmann2006theory}, point to the parameters deeper in the hierarchy having less of an impact on the estimation process. As such, for all present applications I set the $ \delta $ and $ r $ hyperparameters both equal to some small number such as $ 0.1 $. I defer further discussions to the simulation and empirical applications and now elaborate on the asymptotic model diagnostic and Adjusted ROC-Statistic below.

\subsection{Asymptotic Model Diagnostics}\label{sec: asy2}
One of the more useful outcomes of the proposed model in both the nonparametric and parametric implementations is that it simply adds one extra parameter to be estimated. In the parametric case, since we know E($ \alpha^*|\beta $) for existing models such as the logit ($ \alpha^*= 1 $ in the parametric case), we can use large-sample results under i.i.d. assumptions to test the hypothesis that our model results vary from traditional GLM model fits. Indeed the nonparametric methodology is even more useful in this regard. Consider that if $ y=1 $ we have the specified link condition implies, $ \label{estlnkcnd}F(\b \beta|y) = \lambda(X, \b\beta)$. This implies that if we parametrically assume a distribution for $ \hat{F} $ (which we know converges to the true F) such as the Logistic or Probit distribution and input the estimated $ \b{\hat{\beta}} $'s into the functional specification we can calculate the divergence of $ \b{\hat{F}}(\b{\hat{\beta}}|\b y) $ from \eqref{estlnkcnd}. For example, we may compute the value of $ \alpha^* $, say $ \bar{\alpha^*} $ that minimizes,\e
\left(\b{\hat{F}}(\b \beta|\b y)^{\b\alpha^*} - \b\lambda(X, \b\beta)\right).
\ee 
In particular, we know for GLM, if the convergence has occured to the true distribution then $ \alpha^* $ should equal 1. While the X's are held fixed,  $ \bar{\alpha^*} $ is both unbiased, consistent and asymptotically normal by the central limit theorem and i.i.d. assumptions, as long as $ \b{\hat{\beta}} $ is consistent and asymptotically unbiased. The accompanying proofs ensure that this is the case. Accordingly, given $ \b{\bar{\alpha^*}} $, we can thus estimate the aymptotically unbiased and consistent estimates of the variance of $ \alpha^*$ as well to get,
\begin{eqnarray} \alpha^* \sim N(1,E(E(\alpha^*|\b \beta^*)-1|\b \beta^*)^2), \implies \hat{\bar{\alpha^*}} \stackrel{asymp. }{\sim}N\left(1, 
\frac{\sum_{i=1}^n(\alpha_i-1)^2}{n-1}\right). \label{asymNrm}
\end{eqnarray}
$ \b \beta^*$ above represents the optimized estimated value. Thus, we can check our hypothesis that $ \bar{\alpha^*} = 1$ for any particular parametric specification on the probability of success.
\vspace{.1in}
\begin{algorithm}
\hrule
\vspace{.1in}	
\textbf{Large Sample Test for Model Diagnostic.}
\vspace{.1in}
\hrule
\vspace{.1in}
\begin{enumerate}
	\item Perform a t-test on $\hat{\bar{\alpha^*}}$, with the appropriate null hypothesis values, and accept/reject model fit assumptions.
	\item Thus,
	\begin{enumerate}
		\item Under rejection, the existing GLM is not adequate given assumptions on the model specification and the proposed model should be used.
		\item Otherwise, the existing GLM is adequate and it can be used for model fit, inference and prediction (classification) accordingly\footnote{Note however, that model fit, prediction and inference criteria should be evaluated on a wholistic basis to arrive at a choosen model even if the null hypothesis is not rejected.}.
		\newline \hrule
	\end{enumerate}
\end{enumerate}
\end{algorithm}
\noindent This framework can similarly be extended to the likelihood ratio test, under the appropriate null values. Since this algorithm can be directly applied no matter the parametric distribution assumed, without the need to fix any variance parameter of the latent distribution, it extends \cite{liu2018residuals} accordingly. As such, it is applicable to any of the infinitely many distributional assumptions that are possible, without the need to actually make such an assumption in the model specification at all. As such, it is one of the more important contributions of this work.
%

%
\subsection{Asymptotic Distribution of Adjusted ROC-Statistic}

In order to analyze adequacy of classification performance, there are many existing tools such as the Receiver-Operating Curve. Here I consider the Adjusted ROC-Statistic (ARS) instead, based on \cite{chowdhury2019supervised}, as not only does it allow for interpretable estimates, it also has well known closed form large sample distributions. This allows at least two advantages over existing statistics. First, ARS can be represented as a simple interpretable ratio of observed classification outcomes, without the need for a likelihood. Second, the classification performance of any two models can be tested to see if they differ statistically. In addition, the mathematician can employ bootstrap or other methods as well to compare the performance difference between models.
%
%
	
%

Accordingly let, $ G = \text{Ground True} $, $ S(t) = \text{Fitted Prediction Subject to Some Parameter t} $, $ D = \text{Entire Dataset} $, then we may define the following quantities.
\begin{eqnarray}
True \ Positive = \frac{|S(t) \bigcap G|}{|G|},\
True \ Negative = \frac{|\neg S(t) \bigcap \neg G|}{|\neg G|},\\
False \ Positive = \frac{|S(t) - G|}{|D - G|},\
False \ Negative = \frac{|\neg S(t) - G|}{|G|}.
\end{eqnarray}
\vspace{-.2in}
Then, \begin{eqnarray}
ARS = \frac{\frac{|S(t) - G|}{|D - G|}+\frac{|\neg S(t) - G|}{|G|}}{\frac{|S(t) \bigcap G|}{|G|}+\frac{|\neg S(t) \bigcap \neg G|}{|\neg G|}}
\end{eqnarray}
or simply the ratio of incorrectly identified vs. correctly identified elements according to the model fitted. 
%
%

Then ARS can be shown to have an asymptotic distribution \cite{oddsOnline} given by
\begin{eqnarray}
	log (ARS) \sim N(log(Odds ratio),\sigma^2),\
	where\ \sigma=\sqrt{\frac{1}{n_{11}} + \frac{1}{n_{10}} +\frac{1}{n_{01}} +\frac{1}{n_{00}   }  }
\end{eqnarray}
where $ n_{ij} = \text{counts within each cell in the confusion matrix}$. A derivation of this is given in the supplementary materials for this paper. Currently, note that it is entirely plausible that one or more of the cells will be $ 0 $. Thus, to avoid dividing by $ 0 $, I recommend including some small $ \epsilon \ne 0$ for inference. Another approach could be to impose each cell being at least 1 for identifiability.
\subsubsection{Inference}
Clearly, $ ARS \in [0, \infty) $. Then using a slight aberration of the usual hypothesis testing procedure let, 
\begin{eqnarray}
	\nonumber H_0 = \text{Incorrect and correct identification are equally likely}.\\ 
	\nonumber  H_A = \text{Incorrect and correct identification are not equally likely}.
\end{eqnarray}
If incorrect and correct identification are equally likely, then the test statistic becomes, \begin{eqnarray}
	\frac{\sqrt{n}\ log(ARS)}{\sigma}) \sim N(0,1).
\end{eqnarray} 
This framework can be used in a two sample t-test as well to compare any two models fitted to the data. With multiple models, the test can be expanded accordingly. As an example, when the variances for the log-odds attained for ARS for two different samples are assumed to be the same we can do a pooled t-test, 
\begin{eqnarray}
	\text{Test Statistic} =  \frac{\bar{\kappa_1}-\bar{\kappa_2}}{\sqrt{\frac{1}{n_1}+\frac{1}{n_2}}\sqrt{\frac{(n_1-1)s_1^2+(n_2-1)s_2^2}{n_1+n_2-2}}},
\end{eqnarray}
where the subscripts indicate each estimate under the relevant model specifications. If the two models are deemed to be dependent in some manner a matched pair test can be similarly done above. Furthermore, multiple models can be compared using the Wald test. As such, the methodology can be readily extended through a Likelihood Ratio based formulation. Semiparametric estimation of ARS can also proceed straightforwardly and both of these applications are discussed in the supplementary materials. 

\section{Monte Carlo Simulation}\label{sec: mon2}

In order to validate the methodology and the mathematical results, extensive simulation studies were done for both the penalized and unpenalized versions following \cite{CHOWDHURY2021101112}. Various DGPs both symmetric (Logit and Probit) and asymmetric (Complementary Log-Log) were considered, for different sample sizes ($ n=\{100,500,1000, 2000\} $) and models,\begin{eqnarray}
	\b{y} = Intercept + X_1 + X_2,\\
	\b{y} = Intercept + X_1 + exp(X_2),\\
	\b{y} = Intercept + exp(X_1) + sin(X_2).
\end{eqnarray}

Finally, another step was done to create datasets which had different numbers of successes as opposed to failures. Thus, the unbalancedness of the data were varied between $ \{0.1, 0.2, 0.3, 0.4, 0.5\} $. Here, 0.5 indicates equal number of successes and failures (balanced), 0.4 indicates 10\% fewer successes than failures and so forth. Accordingly, for each sample size there are five different unbalanced datasets, each of which has three parameters or $ \beta $'s to estimate for each of the three DGPs for each of the models specified (linear, non-linear or mixed). As such, for each model, there are 60 different datasets, each with 3 parameters to estimate, for a total of $ 180\times 3 = 540 $ parameters to estimate, compare and contrast\footnote{Note also that by construction, we know what the true $ \beta $'s are, and therefore, can use these true values to understand the performance of each of the models fitted to each dataset.}. 

In addition, two separate simulation runs were performed, one in an unpenalized and the other in a penalized formulation. Where for the penalized application a nuisance variable was added to compare model fit and model selection performances concurrently. The results for the unpenalized simulation can be found in Table \ref{simsum1c3}, Table \ref{simsum2c3}, and Table \ref{simsum3c3} below, and the results of the penalized simulation may be found in 
Table \ref{simsum4c3}, Table \ref{simsum5c3}, and Table \ref{simsum6c3} below. The Proposed Unpenalized (not the Bayesian Adaptive Lasso) Nonparametric application uniformly contained the true parameters more often, and thus had better coverage than the other methods compared. It was able to attain this by having confidence interval ranges which were smaller than the Penalized Logistic, which had the worst coverage performance among the methods compared here. Furthermore, this was achieved without considering functional specifications which lack scientific interpretability as in Neural Networks\footnote{This is because deeper networks with more complex basis expansions can lead to scientifically uninterpretable models, at the cost of better classification outcomes.}. 

Accordingly, for Neural Net, since not all model specification and layers could always be fitted, a range of between two to five layers were considered with two neurons in each layer. While a more complicated model could have been used for comparison, the same could be said for the Proposed Nonparametric and Proposed Logistic methods as well. As such, following \cite{CHOWDHURY2021101112} to keep model performance comparable, more complicated model formulations were not deemed appropriate for the NN. In addition, Logistic and Penalized Logistic formulations were not considered for the classification performance comparison given that Chowdhury [Ibid] and \cite{SDSS2021} show that they are outperformed by the other methodologies compared.
\begin{table}[!tph]
	\begin{minipage}{1\textwidth}
		\caption{Simulation Coverage (in Percentage) Summary for Proposed Unpenalized DGPs (at 1\% Significance Level).}
		\begin{center}
				\begin{tabular}{ccccc}\hline\hline
					&Prop. & Prop. & Bys. & Pen.\\
					&   NP. & P.& Prbt. & Logit\\
					\hline
					L. (NL)&98.33\%	& 95.00\%	&83.33\%	 &51.67\% \\
					P. (NL)& 100.00\%&	95.00\% &	75.00\%&46.67\%	\\			
					C. (NL)& 100.00\%&93.33\% &	80.00\% & 56.67\%\\
					L. (Mx.)&98.33\%	& 91.67\%	&71.67\%	 &61.11\% \\
					P. (Mx.)& 100.00\%&	95.00\% &	75.00\%&25.00\%	\\			
					C. (Mx.)& 100.00\%&96.67\% &	81.67\% & 20.00\%\\
					L. (L)&100.00\%	& 96.67\%	& 85.00\%	 & 63.33\% \\
					P. (L)& 98.33\%&	96.67\% &	80.00\%&66.67\%	\\			
					C. (L)& 98.33\%&96.67\% &	81.67\% & 66.67\%\\
					\hline
				\end{tabular}
		\end{center}
		\label{simsum1c3}
	\end{minipage}
	\begin{minipage}{1\textwidth}
			\caption{Simulation Confidence Interval Range Summary for All DGPs (at 1\% Significance Level).}
			\begin{center}
				\begin{tabular}{|cccc}\hline\hline
					Pen.  & Prop.& Prop.  &Bys. \\
					Logistic & NP. & Logit &  Prbt.\\
					\hline
					6.47& 5.66	& 5.37 & 2.00\\
					7.44& 5.42	& 5.20 & 1.77\\
					7.77	& 5.65	&4.89   & 1.88 \\
					7.66	& 	5.75& 5.40   & 2.07\\
					3.94 &	5.64&	5.12 &	1.87\\			
					2.27 & 6.12&	4.93& 1.84	\\
					7.12	& 5.90	& 4.73	 & 1.75  \\
					7.45 &	 5.77&5.15	&1.92	\\			
					6.96&5.73&4.81&1.66\\
					\hline
				\end{tabular}
		\end{center}
		\label{simsum2c3}
	\end{minipage}
	\begin{minipage}{1\textwidth}
		\begin{center}
				\caption{Unpenalized Simulation Summary of ARS for All DGPs Considered.}
				\begin{tabular}{ccccc}\hline\hline
					&Proposed & Proposed& Bayesian &Neural \\
					&   Nonpara. & Logistic& Probit& Net\\
					\hline
					Non-Linear&0.07	& 0.22	& 0.21 &	0.19  \\
					Mixed& 0.11& 0.22&	0.22 & 0.22  \\			
					Linear& 0.08& 0.19 &	0.23 &  0.20		\\
					\hline
					\label{simsum3c3}
				\end{tabular}
		\end{center}
		\scriptsize{Note: This is a summary over all three DGPs, sample sizes and unbalancedness for all models fitted. In total there are 180 parameters per DGP for a total of 540 parameters to be estimated over the entire simulation study. The results are summarized by average over all simulated datasets. L., P., and C. indicate the Logistic, Probit and Complementary Log Log DGPs respectively. NL, Mx, and L indicate Nonlinear, Mixed and Linear DGPs respectively.}
	\end{minipage}
\end{table}
The difference in ARS on average were statistically significant for the Proposed Nonparametric method in comparison to all other models compared for Nonlinear, Mixed and Linear models over all DGPs and datasets considered.

Seaparately, the penalized simulation contained an extra nuisance parameter drawn randomly from the standard normal. Here again, the results were extremely encouraging, and consistent with the results from the previous application. The Proposed Penalized (using Bayesian Lasso) Nonparametric methodology can not only outperform existing models (including the parametric methodology) in inference, but also in classification in regards to ARS. Indeed, the classification results even outperforms the unpenalized (not using Bayesian Lasso) application and Neural Net, on average over all of the many different datasets considered. The inference results again show near perfect coverage results with smaller confidence intervals than the Penalized Logistic\footnote{In the current application, since an extra nuisance variable was considered, the Penalized Logistic model was also considered for comparison.}.
\begin{table}[!tph]
	\begin{minipage}{1\textwidth}
		\caption{Summary of Penalized Simulation \newline Coverage Percentage for All DGPs (at 1\% \newline Significance Level).}
		\begin{center}
				\begin{tabular}{cccccc}\hline\hline
					&Unp. & Pen. & Prop. & Bys. & Pen.\\
					&  NP. & NP. & P. & Prbt.& Logit\\
					\hline
					L.(NL)&100.00	&100.00 & 98.75	&92.50 &70.00 \\
					P. (NL)& 100.00&	98.68&	97.37 &84.21 &65.79	\\			
					C. (NL)& 100.00& 98.75&98.75 &	81.25 & 65.00		\\
					L. (Mx.)&100.00	& 100.00	& 100.00 &81.94	 &70.83 \\
					P. (Mx.)& 97.22 & 97.22 &	94.44 &	81.94 &70.83	\\			
					C. (Mx.)& 100.00 & 100.00 & 96.67 &	81.67 & 72.06		\\
					L. (L)&98.75	& 100.00	& 100.00	 & 88.75& 75.00 \\
					P. (L)& 96.25&	95.00 &	97.50& 80.00 &73.75	\\			
					C. (L)& 100.00&100.00 & 100.00&	87.50 & 75.00\\
					\hline
				\end{tabular}
		\end{center}
		\label{simsum4c3}
	\end{minipage}
	\begin{minipage}{1\textwidth}
		\caption{Penalized Simulation Confidence \newline Interval Summary for All DGPs (at 1\% \newline Significance Level).}
		\begin{center}
				\begin{tabular}{|ccccc}\hline\hline
					Unp.	&Pen.  &  Prop.  &Bys. & Pen. \\
					NP.	&  NP. &  Logit &  Prbt. & Logit\\
					\hline
					5.96& 5.77	& 5.52 & 1.87& 5.64\\
					6.04& 5.60	& 4.93 & 1.99& 5.81\\
					5.88&5.43	& 5.63	&2.03  & 6.18 \\
					5.64	& 	5.87& 5.67   & 1.80 & 6.22\\
					5.67 &	5.78&	5.22 &	1.97 &6.03\\			
					5.97 & 5.44&	5.44& 1.97	& 6.07 \\
					5.74	& 5.66	 & 5.67 & 1.78 & 6.25\\
					5.53&5.15	&4.86	&  1.96& 6.32\\			
					5.74&5.97&5.60& 1.94 &  6.46\\
					\hline
				\end{tabular}
		\end{center}
		\label{simsum5c3}
	\end{minipage}
	\begin{minipage}{1\textwidth}
		\caption{Penalized Application Summary of ARS for All DGPs Compared.}
		\begin{center}
				\begin{tabular}{cccccc}\hline\hline
					&Unpenalized & Penalized & Proposed& Bayesian &Neural \\
					&   Nonpara. & Nonpara&Logistic& Probit& Net\\
					\hline
					Non-Linear	&0.07 	&0.07  & 0.23&	0.25 &0.16 \\
					Mixed& 0.17&0.07 &0.23	 &0.27&0.21  \\			
					Linear& 0.08& 0.05 &	0.19 &  0.23 &0.21		\\
					\hline
				\end{tabular}
		\end{center}
		\label{simsum6c3}
		\scriptsize{Note: This is a summary over all three DGPs, sample sizes and unbalancedness for all models fitted. In total there are 180 parameters per DGP for a total of 540 parameters to be estimated over the entire simulation study. The results are summarized by average over all simulated datasets. L., P., and C. indicate the Logistic, Probit and Complementary Log Log DGPs respectively. NL, Mx, and L indicate Nonlinear, Mixed and Linear DGPs respectively.}
	\end{minipage}
\end{table}
In summary, the results are indicative of the efficiency of the methodology and the mathematical results. The Proposed Nonparametric application contained the true parameters more often than the parametric application, which in turn was more efficient than the other existing methodologies. It did so while having smaller confidence intervals than the Penalized Logistic application. This same superior performance also extended to classification. While the Proposed Parametric application and the existing Bayesian Latent Probit gave classification accuracy similar to Neural Nets, the Proposed Nonparametric applications almost uniformly outperformed all other methodologies on average and were statistically significant in the outperformance. Using these encouraging results I now apply the methodology to real-world dataset applications below and compare its performance to Random Forests, and deep neural networks.

\section{Empirical Application}\label{sec: emp}

I make several empirical applications of the methodology discussed above. The first application is a biomedical one, where we seek to identify intoxicated individuals, based on phone accelerometer data, and the second is an application to identify exotic particles in high-energy Physics. They are detailed below.

\subsection{Detecting Heavy Drinking Events Using Smartphone Data}

\begin{figure}[!htb]
	\begin{minipage}{0.5\textwidth}
		\centering
		\includegraphics[width=0.92\linewidth]{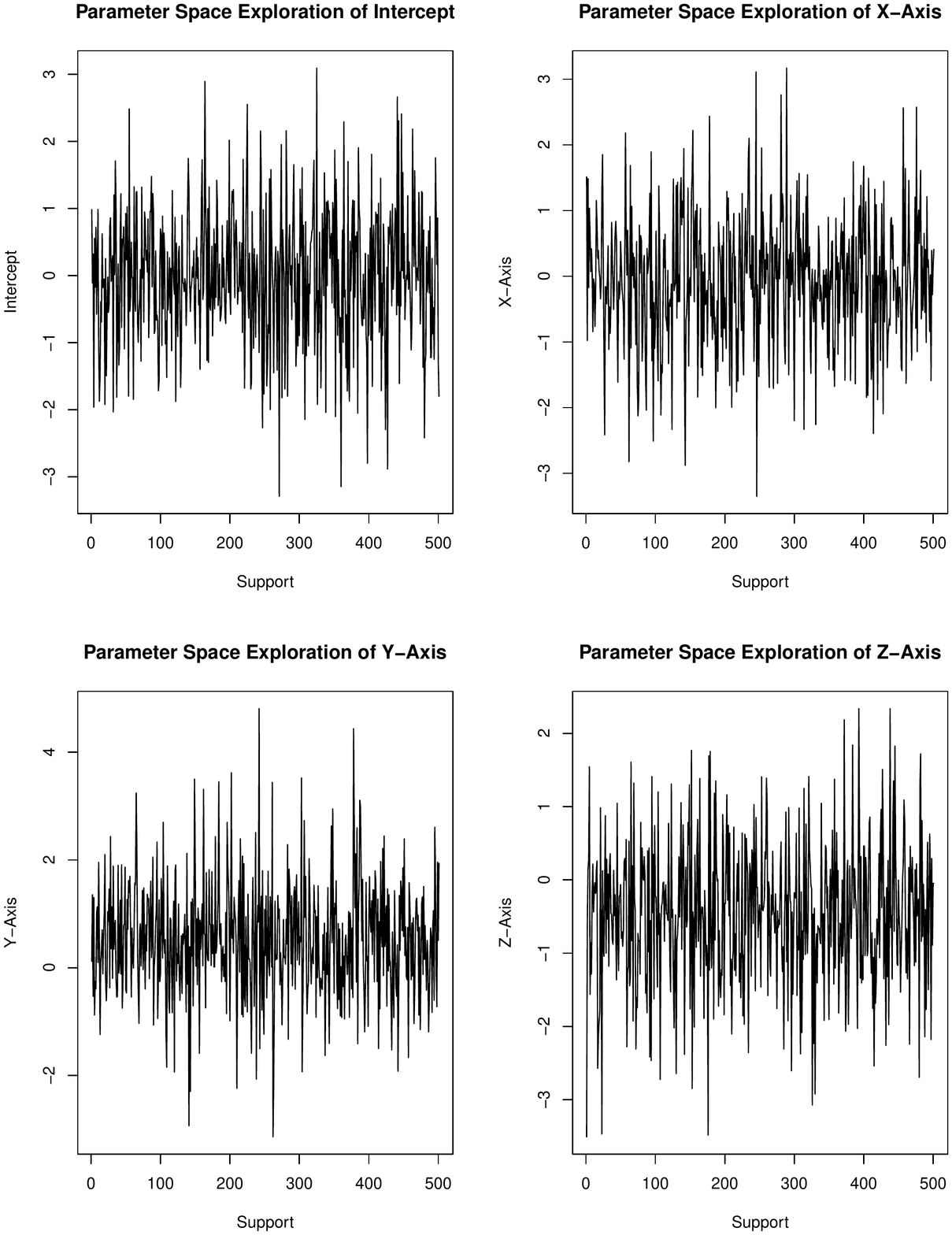}
		\caption{Unpenalized Draws.}
		\label{fig:intxconvc3}
	\end{minipage}%
	\begin{minipage}{0.5\textwidth}
		\centering
		\includegraphics[width=0.92\linewidth]{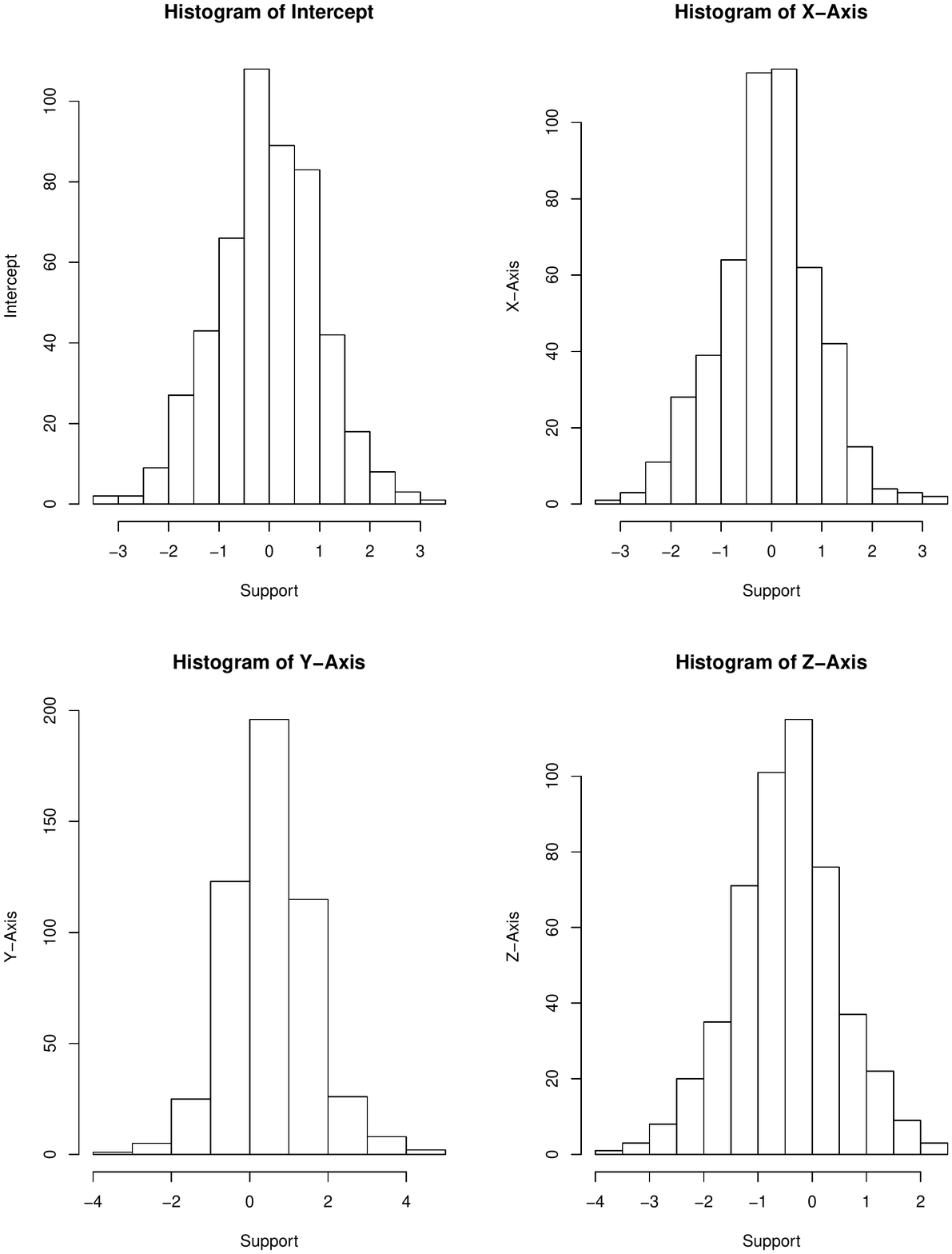}
		\caption{Unpenalized Histograms.}
		\label{fig:intxhistc3}
	\end{minipage}
\end{figure}

To illustrate the efficacy of the model, I apply a simple model specification using its almost sure convergence property, to detect heavy drinking events using smartphone accelerometer data in \cite{killian2019learning}. Given the time series nature of the data the authors identified heavy drinking events within a four second window of their measured variable of Transdermal Alcohol Content (TAC) after various smoothing analyses on the accelerometer data. Their best classifier was a Random Forest with about 77.50\% accuracy. A similar analysis was done on a far simpler model of TAC readings against the accelerometer reading predictors, for all subject's phone placement in 3D space, for the x, y and z axes, \e TAC = Intercept + x-axis\ reading + y-axis\ reading + z-axis\ reading.\ee
TAC here was simply set to 1 if the measurement was over 0.08 and 0 otherwise. The same four second time window of accelerometer readings were used in the analysis with the assumption that the TAC readings were unlikely to change in such a small time interval. The results were extremely encouraging, with Test Data (TeD, 20\% of the data) ARS classification accuracy of nearly 100.00\%, with just $ 1,000 $ iterations and $ 500 $ burn-in period, using some of the methodological contributions in \cite{Dukedissertation21} and \cite{SDSS2021} (the relevant plots can be found in Figure \ref{fig:intxconvc3} and Figure \ref{fig:intxhistc3}).
The strength of the methodology to perform both model fit and model selection at the same time as seen in the simulation studies were also evident here. Specifically, in the penalized application the Adaptive Bayesian Lasso of (\cite{leng2014bayesian}) was applied in a Hierarchical framework. Contrary to the accepted norm that we cannot perform model fit and model selection at the same time, the TeD application had perfect predictive accuracy (the relevant plots for the methodology can be found in the supplementary materials for this article). However, it did have a slightly worse predictive performance in TrD (0.34 vs. 0.30). These new findings are ``significant'' in that they challenge and extend our discussion on scientific and statistical significance considerably. However, I defer further discussions on these findings till the discussion section.

One further advantage of the methodology is that it allows us the ability to perform inference as the almost sure convergence of the parameter estimates retain their interpretability in the current model. Those results can be found in Table \ref{IntxParamC3} below. The results bring to mind the image of a heavily intoxicated individual trying to walk. The nature of the measured data for the Z-axis implies that the Proposed Nonparametric, the Penalized Nonparametric and the Proposed Parametric versions all find only the Z-axis as significant in explaining heavy drinking events. On the other hand the Bayesian Probit finds the Y-axis to be significant. The MLE Logistic and Penalized Logistic both indicated all variables to be significant. Thus, looking simply at the significance criteria, it is not clear which of the methodologies should be relied upon. 
However, when we compare the model fits for the various methodologies, the nonparametric applications stand out as clear winners. The Proposed Nonparametric application had the lowest TeD AIC at 0.94. The next best model fit was for the Proposed Nonparametric Penalized (0.95) application with Proposed Parametric Logistic (0.97) coming in third in this regard. Accordingly, it is clear that in regards to MIPs the Proposed Nonparametric methodologies have a clear advantage in this application over the other existing methods compared.
%
\begin{table}[!tph]
	\caption{Heavy Drinking Event Detection Parameter Summary for All Methodologies.}
	\begin{minipage}{.5\textwidth}
		\begin{center}
				\begin{tabular}{c|cccc}\hline\hline
					Method&Predictor&Est.&CL&CH\\
					\hline
					\multirow{4}{*}{(1)} &	Intercept & $ 0.24^{**} $ & 0.01 & 0.47 \\
					&	X-axis & 0.02 & -0.22 & 0.25\\
					&	Y-axis & 0.03 & -0.19 & 0.25 \\
					&	Z-axis & $ -0.54^{**} $ & -0.83 & -0.26 \\
					\hline
					\multirow{4}{*}{(2)} &				Intercept & 0.22 & -0.03 & 0.48  \\
					&				X-axis & -0.07 & -0.31 & 0.17  \\
					&			Y-axis & 0.21 & -0.04 & 0.46   \\
					&		Z-axis & $ -0.82^{**} $ & -1.05 & -0.59 \\
					\hline
					\multirow{4}{*}{(3)} &	Intercept & -0.13 & -0.3 & 0.05  \\
					&	X-axis & 0.01 & -0.19 & 0.2  \\
					&		Y-axis & 0.07 & -0.13 & 0.27   \\
					&	Z-axis & $ -0.21^{**} $ & -0.37 & -0.05  \\
					\hline
				\end{tabular}
		\end{center}
	\end{minipage}
	\begin{minipage}{.5\textwidth}
		\begin{center}
				\begin{tabular}{c|cccc}\hline\hline
					Method&Predictor&Est.&CL&CH\\
					\hline
					\multirow{4}{*}{(4)}&Intercept & -0.01 & -0.14 & 0.11 \\
					&	X-axis & -0.12 & -0.32 & 0.08  \\
					&	Y-axis & $ 0.24^{**} $ & 0.06 & 0.43  \\
					&	Z-axis & -0.02 & -0.15 & 0.10   \\
					\hline
					\multirow{4}{*}{(5)}&Intercept & $ -0.87^{***} $ & -0.9 & -0.85  \\
					&	X-axis & $ -0.04^{*} $ & -0.09 & 0  \\
					&	Y-axis & $ 0.17^{***} $ & 0.11 & 0.23 \\
					&	Z-axis & $ 0.00^{***} $ & 0.00 & 0.00  \\
					\hline
					\multirow{4}{*}{(6)}&Intercept & $ -0.87^{***} $ & -0.9 & -0.84  \\
					&	X-axis & $ -0.04^{*} $ & -0.11 & 0.02  \\
					&	Y-axis & $ 0.17^{***} $ & 0.09 & 0.25  \\
					&	Z-axis & $ 0.00^{***} $ & 0.00 & 0.00  \\		
					\hline
				\end{tabular}
			\label{IntxParamC3}
		\end{center}
	\end{minipage}
	\begin{minipage}{\textwidth}
		\caption{Heavy Drinking Event Detection ARS Summary}
		\begin{center}
				\begin{tabular}{ccccc}\hline\hline
					&   (1)-TrD.& (1)-TeD.& (2)-TrD.& (2)-TeD.\\
					\hline
					ARS&0.30	& 0.00	& 0.34 &	0.00  \\
					\hline
				\end{tabular}
		\end{center}
		\label{highpartc3}
	\end{minipage}
	\scriptsize{Note: (1) Nonprametric, (2) Penalized Nonparametric, (3) Parametric, (4) Existing Bayesian, (5) MLE Logistic, (6) Penalized Logistic. Est. indicates Estimate. CL and CH indicate Confidence Interval Low and Confidence Interval High respectively. TrD. and TeD. indicate Training and Test Datasets (20\% of the data) respectively} 
\end{table}
\subsection{Exotic Particle Detection Using Particle Accelerator Data}

In order to see the applicability of the methodology across other scientific fields, I now apply the methodology to the identification of high-energy particles in Physics (\cite{baldi2014searching}). There are 28 feature sets in the paper, of which the first 21 features are kinematic properties measured by detectors in the particle accelerator, and the last 7 are high-level features derived from the first 21 to discriminate between the two classes. The classes of 0 and 1 refer to noise and signal, respectively. In addition, the model also incorporates an intercept. \e
Signal/Noise = Intercept + \sum_{i=1}^{28} Feature_i.
\ee
For more information on the actual feature sets I refer the reader to the original paper, and here keep the discussion brief. Further note that, as the last seven features were nonlinear functions of the first 21, the specification remained valid, as inference is not the specific goal here. Given the far larger data size, over the Biostatistics application, I ran LAHEML for $ 5,000 $ iterations with $ 2,500 $ burn-in period. The convergence plots, along with the histograms of each parameter for the unpenalized application may be found below in Figure \ref{fig:higgsconvc3}, Figure \ref{fig:higgshistc3}. The penalized application figures can be found in the supplementary materials for this paper. The penalized and unpenalized estimation formulations were identical to that for the Intoxication application for Biostatistics. Again, the classification outcomes were extremely encouraging, and can be found in Table \ref{highpartc3a} below.
\begin{table}[!tph]
	\caption{Signal/Noise Detection Summary of ARS for Nonparametric Application to Exotic Particle Detection Data.}
	\begin{center}
			\begin{tabular}{ccccc}\hline\hline
				&Unpenalized & Unpenalized& Penalized & Penalized \\
				&   Nonpara. (TrD)& Nonpara. (TeD)& Nonpara. (TrD)& Nonpara. (TeD)\\
				\hline
				ARS&0.36	& 0.06	& 0.44 &	0.14  \\
				\hline
			\end{tabular}
	\end{center}
	\label{highpartc3a}
	\scriptsize{Note: Unpenalized Nonpara. (TrD) is the unpenalized application on the training data, and Unpenalized Nonpara. (TeD) is the unpenalized application on the test data (last 500,000 observations). Penalized Nonpara. (TrD) is the penalized application on the training data, and Penalized Nonpara. (TeD) is the penalized application on the test data (last 500,000 observations).} 
\end{table}
The unpenalized application was especially good for the Test Dataset (TeD), with the penalized version also giving excellent results in TeD, that appear to be an improvement on the initial publication. On average the unpenalized version idetified the correct Signal to Noise almost 79.23\%, of the time, but in TeD it had an accuracy of almost 94.00\%! Accordingly, the efficacy of the model is readily apparent in this application. The penalized application for this dataset did not have better results for the same number of iterations. However, since both formulations were only run for 5,000 iterations it seems plausible that the same pattern seen in the Biostatistics application may also be present here. This is because the penalized version is expected take longer to converge given the extra complexity of the estimation process. 
\begin{figure}[!htb]
	\begin{minipage}{0.5\textwidth}
		\centering
		\includegraphics[width=0.92\linewidth]{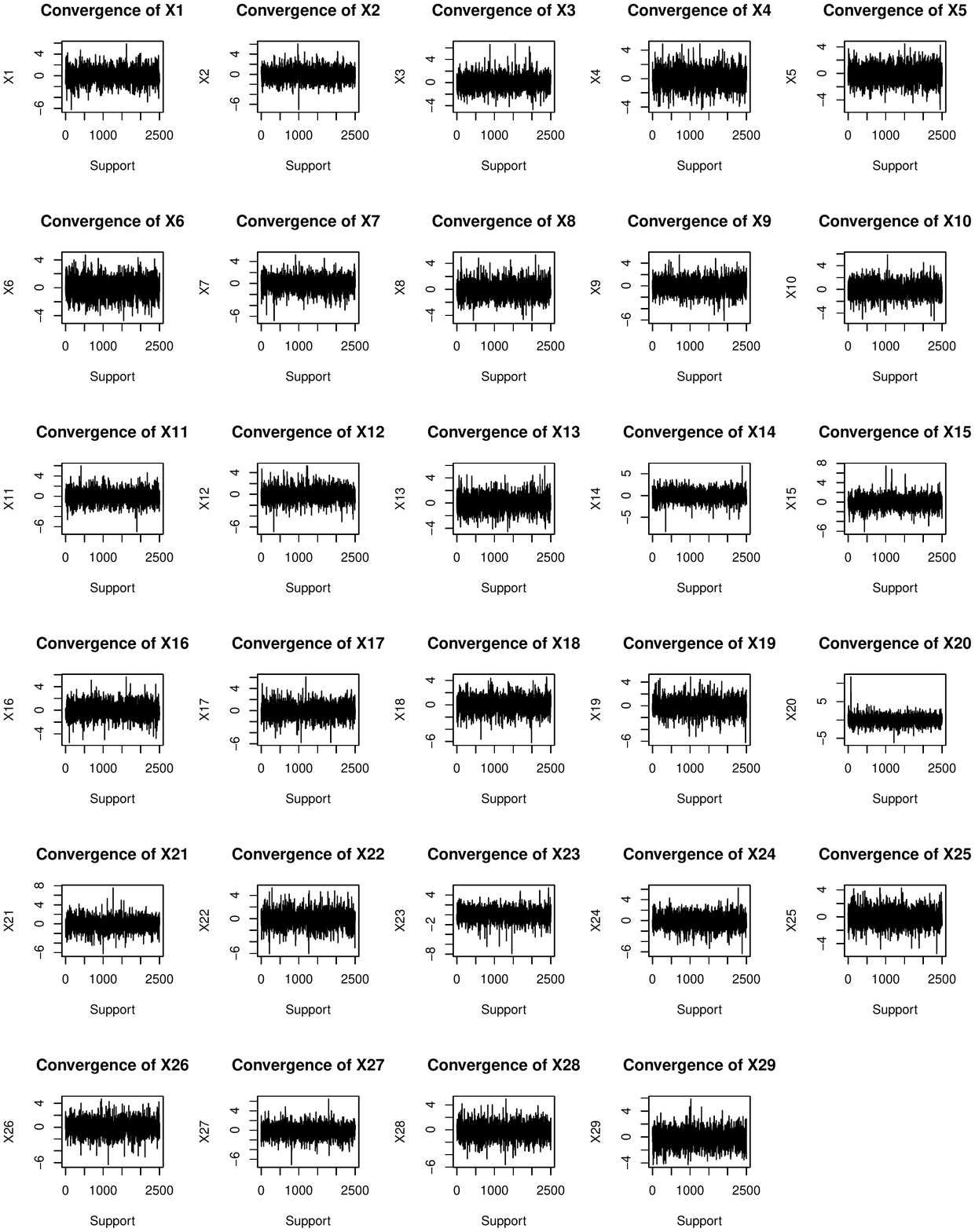}
		\caption{Unpenalized Draws.}
		\label{fig:higgsconvc3}
	\end{minipage}
	\begin{minipage}{0.5\textwidth}
		\centering
		\includegraphics[width=0.92\linewidth]{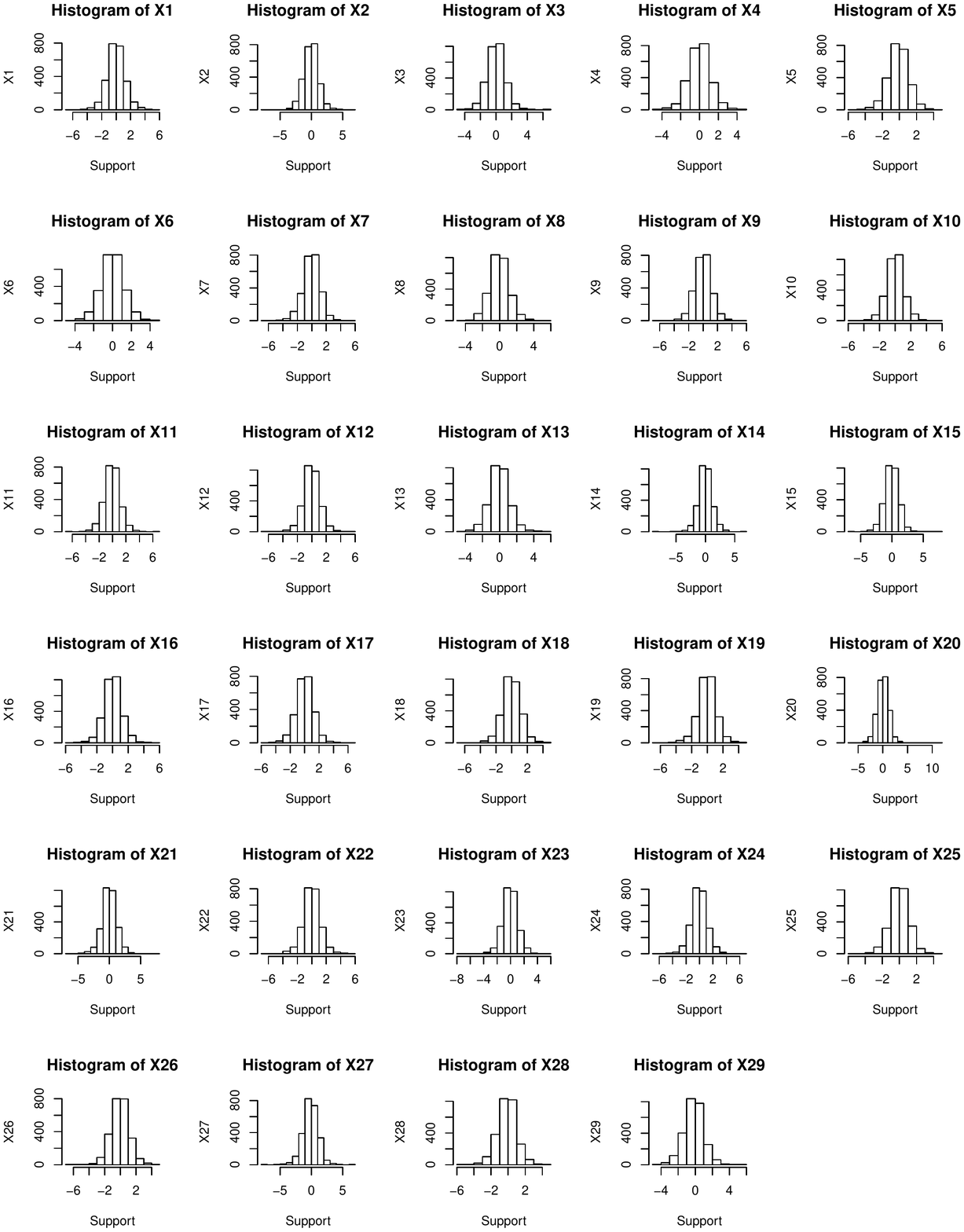}
		\caption{Unpenalized Histograms.}
		\label{fig:higgshistc3}
	\end{minipage}
	\noindent	\scriptsize{Note: The first plot in the upper left corner represents the intercept (X1). All other plots are sequential from left to right as presented in \cite{baldi2014searching}.} 
\end{figure}
\section{Discussion}\label{sec:discch3}
The simulation results are extremely encouraging, because of the variety of datasets on which the different models are compared. Indeed the results validate the theorems and mathematical findings on which they are based. The Proposed Nonparametric method contained the true parameters nearly always with just 5,000 MCMC iterations, even without fixing the variance parameter as is done in existing widely used latent variable formulations. This level of coverage was attained with a smaller confidence interval than the Penalized Logistic regression. Furthermore, the methodology, even in a very simple formulation, easily outperformed ANN for classification, with the difference being statistically significant on average between the two models. Since the interpretation of the parameters remain tractable in the proposed model as opposed to ANN, it further highlights the usefulness of the methodology for myriads of scientific applications.

The application to the real-world datasets also gave extremely encouraging results, yielding deeper insights beyond just the efficiency of the methodology itself. The most apparent of these is that the methodology gives classification results which are 27.10\% better than Random Forests for the biomedical dataset, and had 94.00\% accuracy for the high-energy application in TeD. In addition, the methodology also showed potential for performing model fit and model selection at the same time. The Bayesian Adaptive Lassso application gave results even better than the unpenalized version in the biomedical application, since its classification results were 28.03\% better than the Random Forest application for the dataset. However, for the high-energy particle application this was not the case, with the unpenalized version outperforming the penalized application in both TeD and TrD. This may be explained by the small number of iterations performed, as the extra complexity of the penalized formulation usually requires more iterations.

This highlights the importance of convergence concepts as well as the underlying topological spaces on which they are applied. Stronger convergence coupled with the ability to run it on a stronger topology such as $ L^1 $, means that even simple models can outperform, more complex models on weaker topologies. Further that this may be done without losing scientific interpretability of the parameter estimates. The ability to compare and contrast the suitability of model fits, for any of the infinitely many parametric distributional assumptions also adds another layer of applicability and usefulness to the methodology across the sciences. 

The mathematical results also add to our continuing discussion on the importance of statistical ``significance'' as it relates to scientific significance. They point to the importance of methodologies that have strong convergence of the parameter estimates on stronger topological spaces over weaker convergence concepts (such as convergence in probability or  convergence in distribution) on weaker topological spaces. As such, when inference is of interest, we may proceed using the methodology using simpler and more interpretable models. On the other hand, when classification and/or model fit are the goals, the methodology can be used in conjunction with the many excellent AI and ML models, on stronger topological spaces, for better results accordingly. Therefore, our analytic exercise becomes an attempt to find the best model, using the robust methodology, over finding the significant parameter per se. To be precise, since most models are wrong, but some are useful, the statistical goal can instead focus on robust methodologies, applied in sequentially more complex models, as needed, that rely on scientific interpretability of the model specification. If inference is not the primary goal, then we may improve on the many existing excellent AI and ML methods on stronger topological spaces, to get equivalent yet interpretable results, or in many cases better results as well. This approach gives us a more robust way to correlate scientific and statistical significance concepts to truly give the ``Best of Both Worlds.'' Therefore, there are many possible extensions of the methodology to AI and ML applications across the sciences such as to Neural Networks and Support Vector Machines. However, these concepts require a deeper analysis of the connection between measure spaces and topological spaces, and as such are left for future efforts.

As with any new methodology, however, its true usefulness to the sciences can only be ascertained with broad applications across the sciences, using datasets of varied characteristics. While the mathematical results give solid foundations and explanations for the excellent results, nevertheless, we must be vigilant in its application and estimation. That is, the methodology is extremely versatile in its ability to converge to the true parameter, but this does not preclude the other aspects of good data analysis such as checking for outliers or ensuring the predictors are not correlated with each other etc., especially if inference is the primary goal. However, the simulation results along with the real-world data application outcomes show much potential for the proposed methodology, and further verification is left as an open question to the greater scientific community to explore. 

\section{Conclusion}
\label{sec:conc}
In conclusion, the mathematical foundations and simulation results show the proposed methodology makes notable contributions to widely used methodologies in the sciences. It retains parameter interpretability in a nonparametric setting, while reducing identifiability concerns with near perfect coverage probabilities with smaller confidence intervals than widely used methods. As such, it shows much potential for future real-world data applications. Accordingly, it represents a useful tool for mathematicians, statisticians and scientists to positvely contribute to our continuing conversation on the role of statistical significance and scientific significance and their interplay to answer scientific questions.

\section{Supplementary Materials}
The supplementary materials contain 11 sections. Section I discusses equivalency of binary regression and latent variable formulations and Section II discusses existence and uniqueness of signed measures. Section III provides techinical proofs. Section IV and V provide various specifications of the LAHEML Algorithm. Section VI discusses Asymptotic distribution of ARS. Section VII and VIII gives a Likelihood Ratio test and a Semiparametric algorithm for ARS, respectively. Section IX provides some essential definitions. Section X and XI provide convergence plots and histograms for various data applications.
	
	\bibliographystyle{unsrtnat}
	\bibliography{References}  

		
		
		
		

\end{document}